\documentclass{article} 
\usepackage{iclr2026_conference,times}

\usepackage{amsmath,amsfonts,bm}

\def\eqref#1{equation~\ref{#1}}

\def\1{\bm{1}}

\DeclareMathAlphabet{\mathsfit}{\encodingdefault}{\sfdefault}{m}{sl}
\SetMathAlphabet{\mathsfit}{bold}{\encodingdefault}{\sfdefault}{bx}{n}

\usepackage{hyperref}
\usepackage{url}
\usepackage{booktabs}
\usepackage{graphicx}
\usepackage{amsmath,amssymb}
\usepackage{multirow}
\usepackage{placeins}

\title{Two Axes of LLM Abstention:\\Answer Correctness and Question Answerability}

\author{Benedikt J.\ Wagner\\
City St George's, University of London\\
\texttt{benedikt.wagner@city.ac.uk}
}

\newcommand{\RU}{R_{\mathrm{U}}}
\newcommand{\RW}{R_{\mathrm{W}}}
\newcommand{\CovC}{\mathrm{Cov}_{\mathrm{C}}}
\newcommand{\scoreA}{s_{\mathrm{A}}}
\newcommand{\scoreC}{s_{\mathrm{C}}}
\newcommand{\thrA}{\tau_{\mathrm{A}}}
\newcommand{\thrC}{\tau_{\mathrm{C}}}
\newcommand{\cmark}{\checkmark}
\newcommand{\xmark}{--}

\iclrfinalcopy 
\begin{document}

\maketitle
\lhead{Preprint.}
\thispagestyle{fancy}

\begin{abstract}
Selective answering is usually implemented by thresholding one confidence score. We show
this conflates two failure modes: emitting a wrong answer to an answerable question, and
answering a question that should not be answered at all. Across five instruction-tuned
models from three families (2B--14B), correct-answerable (C), wrong-answerable (W), and
unanswerable (U) questions form a crossed geometry on the same decisions: trained
answerability readouts of output confidence stay at 0.54--0.67 AUROC with no scale
trend, versus 0.97--0.99 from hidden state, while the answerability axis barely
separates C from W, a collapse that persists even when the direction is trained without
any W items. On
naturally occurring false presuppositions (CREPE), where both classes are real user
questions from one distribution, trained output readouts sit at or near chance, elicited self-assessments ($P(\mathrm{IK})$,
$P(\mathrm{True})$, and a direct premise check) reach at best 0.67,
and internal linear readouts 0.69--0.77. The gap is behaviorally consequential and behaviorally repairable: prompting a model to
check premises makes it contest sound and false premises alike (57 percent false
challenges), because it cannot tell which is which; gating the same prompt on the
out-of-sample hidden probe triples challenge precision. We formalize abstention as
three-class selective acceptance with separate budgets on the unanswerable-answer and wrong-answer
rates ($\alpha_U{=}0.15$, $\alpha_W{=}0.50$, $\delta{=}0.10$), certified with exact
binomial bounds on a split independent of score training and threshold tuning. The
certificates expose a scale-dependent asymmetry: the unanswerable-answer budget
certifies on four of five models, while the wrong-answer budget is bounded by model
accuracy; at 8B the factorized two-threshold policy certifies both budgets at 0.75
coverage of correct answers (a single answer-confidence threshold: 0.31; a learned
two-signal fusion reaches 0.81 but forfeits the per-axis attribution of risk), and at
14B it is the only policy that certifies at all. Prior work established that answerability is
internally legible and that pre-generation knowledge can be estimated; our contribution
is the same-decision decomposition of the two axes, its measurement under controls that
bound what output confidence, elicitation, and surface features each recover, and its
risk-controlled composition.
\end{abstract}

\section{Introduction}

A model that abstains well must get two different things right. It must avoid emitting
answers that are wrong, and it must avoid answering questions that should not be answered
at all, such as unanswerable questions and false premises. The dominant abstention recipe
estimates a single quantity, the uncertainty of the generated answer, and thresholds it:
semantic entropy \citep{farquhar2024detecting}, conformal abstention
\citep{yadkori2024mitigating}, maximum-probability scoring \citep{hendrycks2017baseline}.
A single threshold reads low support as rejection, an implicitly closed-world rule; the
questions that should not be attempted at all form an open region that this rule cannot
represent, in the sense of open-world versus closed-world reasoning
\citep{wagner2022neural}.

We show these two requirements correspond to two axes that separate on the same decisions
and are not captured by the same signal. On the SelfAware benchmark \citep{yin2023large}
we score each question once and label three states: correct-answerable (C),
wrong-answerable (W), and unanswerable (U). Ordinary answer-confidence orders them C
above W and U, but barely separates W from U: it mostly tells you whether an attempted
answer succeeded, not whether the attempt was admissible. The answerability axis,
separating answerable from unanswerable, requires a different signal. The pattern
is not a small-model artifact: it replicates across Gemma~2~2B, Qwen2.5-3B, Qwen2.5-7B,
Llama-3.1-8B, and Qwen2.5-14B, and it carries over to naturally occurring false
presuppositions from CREPE \citep{yu2023crepe}, where both question classes are real
user questions from the same source and time period.

We are deliberate about novelty, because the nearest prior work is close. That hidden
states encode unanswerability despite overconfident generation was shown by
\citet{slobodkin2023curious}; linear answerability directions that transfer to SelfAware
and false-premise data were given by \citet{lavi2026detecting}; the value of a
pre-generation, query-level uncertainty signal distinct from answer-level uncertainty was
argued in concurrent work \citep{chen2026query}; and that models carry both a
$P(\mathrm{True})$ and a $P(\mathrm{IK})$ signal is due to \citet{kadavath2022language}.
We do not claim to discover that the answerability axis is internally legible. Our
contributions are: (i) the crossed, same-item geometry of C, W, and U under two signals,
replicated from 2B to 8B across three model families, including a check that the
internal C--W collapse is not an artifact of the probe's training labels; (ii) honest
controls showing the answerability signal is recoverable by surface and elicitation as
well as by a probe, which narrows the claim about output confidence to ordinary
answer-likelihood, together with a natural-data test on CREPE where the surface
confound is far weaker by construction; (iii) a behavioral consequence and its repair: instructed premise-checking fails
because the model contests indiscriminately, while routing the same instruction with the
hidden probe recovers premise correction at a fraction of the false-challenge cost; and
(iv) a formalization of abstention as three-class selective acceptance with separately
certified per-axis risk budgets, whose dual-risk certificate turns a vague capability
claim (``the model is too weak to bound wrong answers'') into a measurable,
scale-dependent quantity, together with a seed-replicated audit of how often issued
certificates hold on fresh splits.

\section{Related work}

\citet{slobodkin2023curious} are the nearest predecessor: they show that models generate
overconfident answers to unanswerable questions while their hidden states encode
answerability, and they causally erase the answerability subspace. Our output geometry is
consistent with their overconfidence finding; our addition is the crossed three-state
treatment on the same items and the certified two-signal policy.
\citet{lavi2026detecting} identify linear unanswerability directions in two open-weight
models across four extractive-QA datasets, transfer them to SelfAware and CREPE, and
steer abstention causally; our answerability direction is the same kind of object, and we
do not claim it as novel. Query-level uncertainty work \citep{chen2026query} explicitly
separates whether the answer is right from whether the query is answerable and estimates
the latter before generation; our distinction is the controlled same-item geometry rather
than introducing the query-level signal. \citet{kadavath2022language} distinguish
$P(\mathrm{True})$ from $P(\mathrm{IK})$; we run both as first-class output baselines,
alongside a direct premise-check elicitation and trained readouts of answer-confidence
features. Sampling-based semantic entropy \citep{farquhar2024detecting} does not fit our
single-pass protocol and is not run here; conformal abstention
\citep{yadkori2024mitigating,mohri2024language} appears as the calibrated
answer-confidence-only policy in Section~\ref{sec:policy}. Closest on the policy side,
\citet{phillips2026entropy} show that entropy alone is insufficient for safe selective
prediction and combine it with a correctness probe; they fuse two signals to detect one
outcome, answer incorrectness, whereas we decompose selective failure into
wrong-answerable and unanswerable states and calibrate a separate risk budget for each.
Internal-state truth and uncertainty probing
\citep{azaria2023internal,chen2024inside,kossen2024semantic} and uncertainty
disentanglement \citep{mucsanyi2024benchmarking} establish that uncertainty is
multidimensional; concurrent work shows uncertainty and correctness are encoded by
functionally distinct internal features \citep{patel2026uncertainty}, which supports the
premise of factorization but builds no abstention policy. Our contribution is the
specific answerability-versus-correctness decomposition in generation and its certified
use, not the general claim.

Finally, the decomposition itself has a neuro-symbolic reading. Thresholding one
confidence score enforces an implicitly closed-world treatment of support: low support
is read as rejection. Three-valued neuro-symbolic reasoning instead makes the unknown
region a first-class object: \citet{wagner2022neural} construct an explicit
open/closed-world boundary, with a negative-complement region separating what a system
can decide from what lies outside its knowledge, in the Logic Tensor Networks framework
of \citet{badreddine2022logic}, and grounding named concepts in learned representations
is the neuro-symbolic notion of conceptual grounding \citep{wagner2021interactive}. Our
C/W/U geometry is an empirical LLM instantiation of that boundary: the admissibility
axis separates the model's closed fragment, where attempts succeed or fail, from its
open region, where attempting is the error; and the factorized policy of
Section~\ref{sec:policy} is a thresholded conjunction of two named predicates,
Answerable and Correct, grounded in hidden states.

\section{Two axes on the same decisions}
\label{sec:geometry}

\paragraph{Setup.} We use SelfAware \citep{yin2023large}, scoring a balanced sample per
model (150 answerable and 150 unanswerable; 120 each for Qwen2.5-3B) once per question,
on five instruction-tuned models from three families: Gemma~2~2B-it, Qwen2.5-3B,
Qwen2.5-7B, Llama-3.1-8B, and Qwen2.5-14B. The 2B and 3B models run in full precision;
the 7B and 8B models run 8-bit quantized and the 14B model 4-bit, and we validated the
quantized pipeline against the full-precision one on Qwen2.5-3B (identical generation
accuracy, answerability readout 0.972 vs 0.973). From the same forward pass we record the output answer-confidence (mean
token log-probability of the greedy answer, and first-token predictive entropy) and the
final prompt-token hidden states at three depths. Correctness on answerable items is
judged by matching the generated answer against the benchmark's gold aliases. SelfAware
answerable accuracy is 0.28, 0.30, 0.35, 0.42, and 0.37 respectively, and we report
exact denominators and treat the correctness axis with appropriate caution.

\paragraph{Symmetric readouts.} To avoid comparing a trained probe against an untrained
scalar, we train equal-capacity $L_2$-regularized logistic readouts for all four
source-by-axis cells, on the same calibration and test partitions, with the hidden layer
chosen on a nested calibration split. Output features are the two answer-confidence
statistics; hidden features are the prompt-token state. Percentile-bootstrap 95\% intervals over test items are in brackets.

\begin{table}[t]
\centering
\caption{Symmetric $2\times2$ readouts (AUROC, test half; bootstrap 95\% CIs). O =
output confidence features, H = hidden state; C = correctness axis, A = answerability
axis. The output-to-answerability deficit persists at every scale with no trend
(O$\to$A 0.54--0.67; at 14B it is smallest, 0.54); the hidden correctness readout
grows with scale.}
\label{tab:2x2}
\small
\resizebox{\linewidth}{!}{\begin{tabular}{lccccc}
\toprule
Source $\to$ axis & Gemma 2 2B-it & Qwen2.5-3B & Qwen2.5-7B & Llama-3.1-8B & Qwen2.5-14B \\
\midrule
Output $\to$ correctness   & 0.82 [0.71, 0.93] & 0.80 [0.66, 0.91] & 0.77 [0.64, 0.89] & 0.84 [0.74, 0.92] & 0.64 [0.51, 0.76] \\
Output $\to$ answerability   & 0.56 [0.46, 0.65] & 0.62 [0.52, 0.72] & 0.64 [0.55, 0.73] & 0.67 [0.59, 0.76] & 0.54 [0.46, 0.64] \\
Hidden $\to$ correctness   & 0.64 [0.49, 0.76] & 0.59 [0.42, 0.75] & 0.80 [0.69, 0.90] & 0.88 [0.79, 0.94] & 0.86 [0.77, 0.94] \\
Hidden $\to$ answerability   & 0.98 [0.95, 1.00] & 0.97 [0.93, 1.00] & 0.98 [0.96, 1.00] & 0.99 [0.97, 1.00] & 0.99 [0.97, 1.00] \\
\bottomrule
\end{tabular}
}
\end{table}

Two points follow, and we state both honestly. First, hidden state reads correctness
moderately at 2B--3B (0.64 and 0.59) and increasingly well at 7B--8B (0.80 and 0.88), so
we do not claim internal state lacks a correctness signal; prompt-time states are known
to predict correctness \citep{morenocencerrado2025noanswer,patel2026uncertainty}, and at
8B the hidden readout is the better correctness reader. What does not change with scale
is the answerability column: even a trained output readout barely reads answerability
(0.56 to 0.67 across all four models), while the answerability axis is read at 0.97 to
0.99 from hidden state. The crossed pattern that survives every model we test is that
ordinary answer-confidence reads execution (did the attempt succeed) and remains far
behind on admissibility (should the attempt have been made). This is a crossed score
geometry, not a strict double dissociation, and we present it as such.

\paragraph{The three-state geometry.} The cleanest statement is geometric. In
calibration-standardized units, ordinary answer-confidence has means $C=0.77$, $W=-0.23$,
$U=-0.05$ on Gemma, and the answerability readout has means $C=0.90$, $W=0.86$,
$U=-0.90$. The pattern is an L shape. Answer-confidence separates C from W and U ($C-W$
gap 1.00, interval $[0.73, 1.24]$) but does not separate W from U (difference $-0.19$,
interval $[-0.43, 0.06]$, including zero). The answerability readout does the opposite,
separating W from U (difference 1.76, interval $[1.67, 1.86]$) while not separating C
from W (difference 0.04, interval $[-0.07, 0.15]$, including zero). Qwen shows the same L
shape with one caveat we report rather than smooth over: its internal C-versus-W collapse
is equally clean (difference 0.01, interval $[-0.19, 0.18]$), but its output W-versus-U
difference is small yet statistically nonzero ($-0.32$, interval $[-0.61, -0.03]$). The
same L shape holds at scale: on Qwen2.5-7B the output W-versus-U difference is 0.03
(interval $[-0.27, 0.30]$) against a C-versus-W separation of 0.82, on Llama-3.1-8B the
figures are 0.14 $[-0.11, 0.39]$ against 1.03, and on Qwen2.5-14B they are $-0.16$
$[-0.44, 0.11]$ against 0.82; the internal W-versus-U separations there are 1.73, 1.67,
and 1.77, while the internal C-versus-W differences stay far smaller (0.09, 0.12, and
0.04). We do not read intervals containing zero
as proof of no separation; the quantitative statement is a strong asymmetry of gap
ratios, reported in full in Table~\ref{tab:gaps} (Appendix~\ref{app:layers}): the
output W-versus-U gap is at most 0.35 of its C-versus-W gap (range 0.03--0.35 across the
four models), and the internal C-versus-W gap is at most 0.07 of its W-versus-U gap
(range 0.00--0.07). Confidence differentiates successful from unsuccessful execution
while barely distinguishing inadmissible from merely unsuccessful attempts; the
answerability axis does the opposite. All intervals here are 95\% percentile bootstrap
over test items (items are the resampling unit); with many contrasts reported, we rest
no single claim on one interval.

\paragraph{The collapse is not an artifact of the probe's labels.} Because the
answerability readout is trained on answerable-versus-unanswerable labels, under which C
and W share a label, its C$\sim$W collapse could in principle be built in by the
training objective. It is not: training the direction on C-versus-U items only, with all
W items excluded from training, still places held-out W items with C rather than U on
every model (W-versus-U AUROC 0.97--0.98; W sits 0.71--0.85 of the way from the U mean
to the C mean; Table~\ref{tab:cvu} in the appendix). Since correctness is decodable
from the same states (H$\to$C up to 0.88), the direction could have absorbed
correctness and separated W from C; it does so only mildly (C-versus-W 0.60--0.70). The
internal representation itself, not the label scheme, groups wrong-but-answerable with
correct-answerable and apart from unanswerable.

\begin{figure}[t]
\centering
\includegraphics[width=\linewidth]{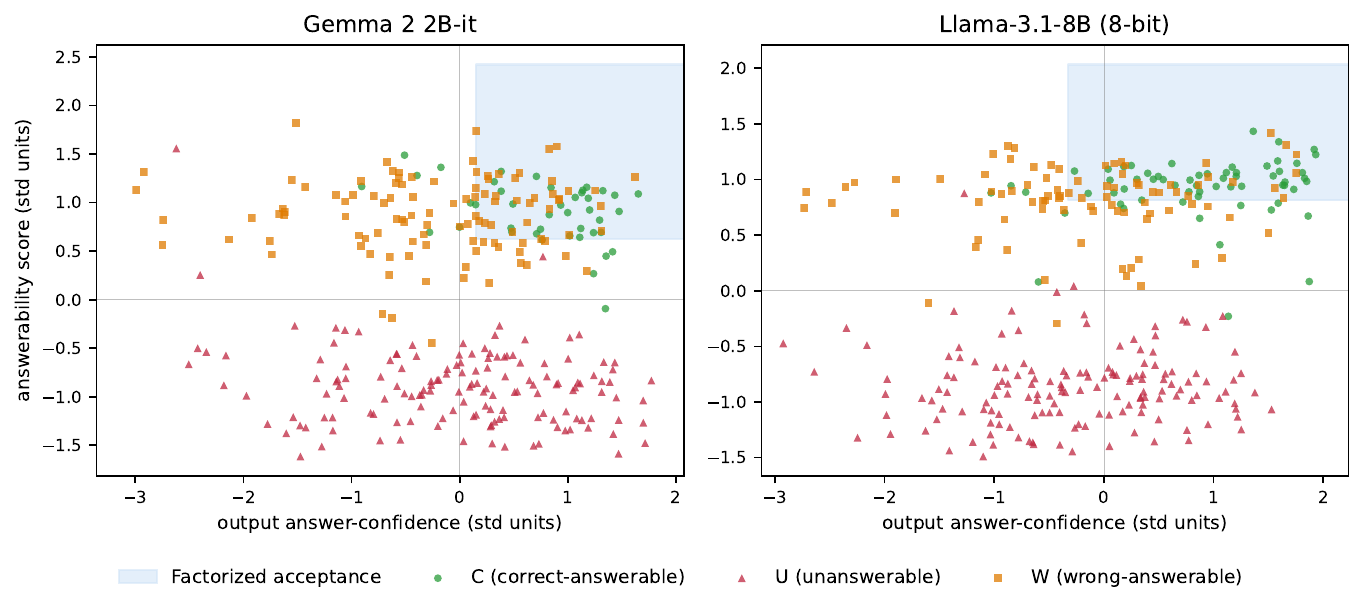}
\caption{The crossed three-state geometry at 2B (left, full precision) and 8B (right,
8-bit). Correct-answerable (C), wrong-answerable (W), and unanswerable (U) questions in
the plane of output answer-confidence (horizontal) and answerability score (vertical).
Answer-confidence separates C from W and U; the answerability score separates U from C
and W. The shaded corner is the Factorized Abstention acceptance region, where both
thresholds are cleared. Axes clipped at the 0.5th confidence percentile for display.}
\label{fig:geometry}
\end{figure}

\section{What the answerability signal is, and is not}
\label{sec:controls}

A high hidden-probe AUROC on SelfAware does not establish that answerability is uniquely
or non-trivially internal, and we ran the controls the claim requires.

\paragraph{Surface confound.} SelfAware draws answerable and unanswerable questions from
different sources and semantic categories, so they differ lexically. A hashed
bag-of-words logistic on the question text alone predicts answerability at AUROC 0.87
(a property of the question sample, independent of any model), close to the hidden
probe. Most of the SelfAware answerability signal is therefore available at the surface,
and a probe's 0.97--0.99 cannot be read as a clean internal representation of
answerability on this benchmark.

\paragraph{Elicitation.} Answerability is also output-visible if you ask for it. An
explicit ``can you correctly answer this question'' elicitation, $P(\mathrm{IK})$ in the
sense of \citet{kadavath2022language}, reaches AUROC 0.88 on Gemma 2 2B-it (mean
$P(\mathrm{IK})$ 0.77 on answerable, 0.24 on unanswerable) and 0.81 on Llama-3.1-8B,
but only 0.61--0.66 on the Qwen models (Appendix~\ref{app:controls}): elicitation can
recover answerability, but how much is model-dependent, and it never matches the hidden
readout. So output is not blind to answerability in
general; what is blind is ordinary answer-likelihood (0.56--0.67 even as a trained
readout). The correct statement is that the dominant recipe, thresholding
answer-confidence, misses answerability, not that no output signal can recover it, on
this benchmark; on naturally occurring false premises even direct elicitation recovers
much less (Section~\ref{sec:crepe}).

\paragraph{Beyond surface.} The component of the internal direction that is not surface
lexicality is shown by transfer across surface-disjoint constructions. A matched-template
set asks ordinary factual questions about nonce, non-existent countries (unanswerable)
versus real prominent countries (answerable); its unanswerable questions share no
vocabulary or genre with SelfAware's philosophical ones. A probe trained on SelfAware
scores this set's answerability at 0.76, and the reverse transfer is 0.64 (Gemma 2
2B-it; this control is single-model). The same probe
transfers to CREPE's natural false presuppositions at 0.65--0.67
(Section~\ref{sec:crepe}), close to the within-CREPE readout, evidence of a shared
answerability component that is not reducible to any one construction's surface. On the
nonce set output confidence is not blind (0.90), because the model hesitates on entities
it recognizes as fictional; the output blind spot is specific to the
confident-confabulation regime, real-world unanswerable and false-premise questions the
model answers without hesitation, which is exactly the CREPE regime where the trained
output readout sits at chance.

We therefore do not call this section a confound that is ruled out. It is a set of
controls that bound the claim: the answerability axis is real and partly internal, but on
SelfAware it is largely surface-available, and our defensible novelty is not its internal
legibility.

\section{Naturally occurring false presuppositions: CREPE}
\label{sec:crepe}

SelfAware's unanswerable questions differ from its answerable ones in genre and
vocabulary, a confound Section~\ref{sec:controls} quantified. CREPE
\citep{yu2023crepe} removes that confound by construction: all questions are real user
questions from the same source (Reddit ELI5) and consecutive time periods, about a
quarter of which rest on a false presupposition, as annotated in the benchmark. A model
that answers such a question fluently commits exactly the admissibility failure the
answerability axis is meant to catch. This is the natural-data test of the geometry: no
templates, no constructed nonce entities, both classes drawn from one distribution. We
score a balanced sample of 500 normal and 500 false-presupposition questions from the
official test split under the identical protocol (Appendix~\ref{app:prompts}); because
ELI5 answers are long-form, CREPE contributes the admissibility axis only.

\begin{table}[t]
\centering
\caption{CREPE: detecting naturally occurring false presuppositions (admissibility
AUROC, test half). Raw = answer-confidence as used in practice; readouts are
equal-capacity logistics; the difference-of-means direction is the training-light
alternative of \citet{lavi2026detecting}. Elicitations ask the model directly
(Appendix~\ref{app:prompts}); the transfer row applies each model's SelfAware-trained
probe to CREPE.}
\label{tab:crepe}
\small
\resizebox{\linewidth}{!}{\begin{tabular}{lccccc}
\toprule
Signal & Gemma 2 2B-it & Qwen2.5-3B & Qwen2.5-7B & Llama-3.1-8B & Qwen2.5-14B \\
\midrule
Raw answer-confidence & 0.49 & 0.55 & 0.52 & 0.53 & 0.58 \\
Output readout (trained) & 0.50 & 0.54 & 0.50 & 0.53 & 0.58 \\
$P(\mathrm{IK})$ elicitation & 0.66 & 0.59 & 0.60 & 0.63 & 0.54 \\
$P(\mathrm{True})$ self-evaluation & 0.54 & 0.58 & 0.55 & 0.54 & 0.55 \\
Premise-check elicitation & 0.66 & 0.43 & 0.64 & 0.65 & 0.67 \\
Bag-of-words (surface bound) & 0.59 & 0.59 & 0.59 & 0.59 & 0.59 \\
Hidden readout (trained) & 0.70 & 0.69 & 0.71 & 0.73 & 0.71 \\
Difference-of-means direction & 0.74 & 0.76 & 0.77 & 0.74 & 0.78 \\
\midrule
SelfAware$\to$CREPE transfer & 0.65 & 0.65 & 0.66 & 0.67 & 0.70 \\
\bottomrule
\end{tabular}
}
\end{table}

Three findings, stated plainly. First, the output blind spot is sharper on natural data
than on SelfAware: trained readouts of answer-confidence features detect false
presuppositions at or never far from chance on all five models (AUROC 0.50--0.58; raw
confidence 0.49--0.58). The models answer questions with false premises exactly as fluently as
sound ones. Second, the internal signal survives the removal of the surface confound but
shrinks: hidden logistic readouts reach 0.69--0.73 and difference-of-means directions
0.74--0.78, against a bag-of-words bound of 0.59 (a property of the shared question
sample, identical for all models, versus 0.87 on SelfAware). Notably, the elicitations that
recover answerability on SelfAware do not recover false-premise detection: $P(\mathrm{IK})$
reaches only 0.54--0.66, post-generation $P(\mathrm{True})$ self-evaluation 0.54--0.58,
and the direct premise-check 0.64--0.67 on four models and below chance (0.43) on
Qwen2.5-3B, which affirms nearly every question as sound; every output-side signal falls
below every within-CREPE internal readout. Asked outright, even the best models
mostly assert their questions are sound (mean premise-check scores 0.93 on sound versus
0.84 on false-premise questions for Qwen2.5-7B). On natural data, no output-side signal
we test matches the internal one. Most of SelfAware's 0.97--0.99
separability is benchmark-specific; the transferable core is moderate. The
SelfAware-trained probe transfers to CREPE at 0.65--0.70 on every model, close to the
within-CREPE readout, so the constructions share one answerability direction, consistent
with \citet{lavi2026detecting}. What matters for policy is that the asymmetry persists
undiminished: internal-minus-output is $+0.15$ to $+0.21$ on natural data, and only one
side of it is usable. Third, the usable side supports risk control at a price. With a
leak-free four-way split (probe-train, tune, certify, test), a frozen admissibility
threshold certifies a false-premise answer rate of at most $\alpha_{FP}=0.15$ on four
of five models, with held-out test rates of 0.07--0.09 and admissible coverage from
0.09 (3B) to 0.23--0.25 (7--8B; Table~\ref{tab:crepegate}); on Qwen2.5-14B the
certification-split bound narrowly fails (8/77, UCB 0.18) while its held-out test rate
is 0.066, within target, consistent with the replication audit's finding that
certificates issue in about half of resplits. We quote certificates against the target rather than the
sometimes-zero-count certification bound, whose point value is fragile: on Llama the
certification split accepted zero of 77 false-premise items (UCB 0.038) while the test
split showed 7 of 91, still well inside the target. A replication audit over 100 random
resplits per model re-runs the whole pipeline (split, probe training, tuning,
certification) end to end: certificates are issued in 45 to 55 of 100 resplits per
model, and issued certificates are violated on the held-out test split in 0 to 3.8
percent of cases, within the 5 percent allowance of $\delta{=}0.05$
(Appendix~\ref{app:replication}). Certified natural-false-premise gating is possible
but expensive with a 0.7 signal; with an at-chance signal it is impossible at any
coverage.

\section{From detection to repair: probe-routed premise checking}
\label{sec:repair}

The admissibility gap is not only measurable but behaviorally consequential, and the
probe can close part of it. We test this with a pre-registered pilot on Llama-3.1-8B
(hypotheses, kill-criteria, and design fixed in the released script before running):
120 false-premise and 120 sound CREPE questions, drawn exclusively from the half of the
sample the admissibility probe was not trained on, under two prompts. The standard
prompt asks for an answer; the challenge prompt adds ``if the question contains a false
assumption, briefly point out what is wrong with it instead of answering.''

\textbf{Instructed premise-checking fails for the reason the elicitation numbers
predict.} Under the standard prompt the model explicitly contests 0\% of false premises
(strict template detector): it answers them fluently. Under the challenge prompt it
contests 74\% of false premises, but also 57\% of sound questions, inventing
premise-denials for questions with nothing wrong (``it is not mandatory to have
headlights on during the day''; ``Neosporin is not an antibiotic''). The instruction
moves the model's threshold, not its knowledge: unable to tell which premises are false
(premise-check elicitation 0.65, Section~\ref{sec:crepe}), it complies by contesting
indiscriminately. This is consistent with the finding that intrinsic self-correction
fails without an external signal \citep{huang2024large}.

\textbf{The probe supplies the discrimination the prompt lacks.} Routing between the
two prompts with the out-of-sample hidden probe (challenge only when the admissibility
score is low) restores selectivity. At a threshold routing 35\% of questions, the
probe-gated policy contests 42\% of false premises at a 14\% false-challenge rate,
against 26\% at 20\% for random gating with the same budget and 74\% at 57\% for
always-challenge: the probe roughly triples challenge precision
(Table~\ref{tab:router}).

\textbf{Correction quality is the remaining bottleneck, reported in layers.} Contesting
a false premise helps only if the correction is right. Scoring the challenge outputs
against CREPE's gold corrections and annotated presuppositions with NLI, 47\% (Llama)
and 61\% (Qwen) of false-premise items receive an NLI-validated correct contest (on
Llama, 39\% are both explicit and validated; 53\% of its explicit contests validate). Some unvalidated contests are wrong
corrections confidently stated, the same confabulation reappearing one level up; others
are NLI misses on paraphrase or annotation ambiguity, so the truth lies between the
strict and lenient bounds and a manual audit sample is released with the code. The
pattern replicates on Qwen2.5-7B in a starker form: under the challenge prompt it
contests 76\% of false premises and 78\% of sound questions, no discrimination at all,
while probe-gating at the same budget yields 41\% against 16\%
(Appendix~\ref{app:router}). Both pilots use $n{=}120$ per class.

\begin{table}[t]
\centering
\caption{Probe-routed premise checking on CREPE (strict premise-contest detector,
held-out items, routing budget 35\%). Cells are FP-contested / sound-falsely-contested
rates. Random gating at the same budget shows the routing value is the probe's
discrimination, not the budget.}
\label{tab:router}
\small
\begin{tabular}{lcc}
\toprule
Policy (FP / sound contested) & Llama-3.1-8B & Qwen2.5-7B \\
\midrule
Never challenge & 0.00 / 0.00 & 0.00 / 0.00 \\
Always challenge & 0.74 / 0.57 & 0.76 / 0.78 \\
Random gate & 0.26 / 0.20 & 0.27 / 0.27 \\
Probe-gated & 0.42 / 0.14 & 0.41 / 0.16 \\
\bottomrule
\end{tabular}

\end{table}

\section{Factorized abstention and dual-risk certification}
\label{sec:policy}

We formalize abstention as three-class selective acceptance. With C, W, and U the
correct-answerable, wrong-answerable, and unanswerable items and a policy $\pi$ that
emits or abstains, define $\RU = P[\mathrm{emit} \mid U]$, $\RW = P[\mathrm{emit} \mid
W]$, and $\CovC = P[\mathrm{emit} \mid C]$. The goal is to maximize $\CovC$ subject to
separate budgets on the two error rates, $\mathrm{UCB}(\RU) \le \alpha_U$ and
$\mathrm{UCB}(\RW) \le \alpha_W$. This is a three-valued decision in the open-world
sense \citep{wagner2022neural}: U is not ``probably wrong'' but ``outside the closed
fragment,'' and the policy composes two $[0,1]$-valued grounded predicates by a
thresholded conjunction, the simplest of the logical compositions studied in
neuro-symbolic frameworks \citep{badreddine2022logic}.

Neither axis-specific score alone is sufficient. A calibrated threshold on
answer-confidence $\scoreC$, i.e.\ conformal abstention on the confidence score in the
sense of \citet{yadkori2024mitigating}, cannot reject U, because W and U overlap on
$\scoreC$; to hold $\RU$ it must over-abstain,
and its certified C-coverage stays at 0.12--0.31 where it certifies at all (its test
coverage never exceeds 0.41 on any model, and at 14B it fails certification outright,
accepting too many unanswerable items). A threshold on the answerability score
$\scoreA$ cannot bound wrong answers below the model's own wrong-rate among admissible
attempts, because C and W largely overlap on $\scoreA$: its certification-split
$\mathrm{UCB}(\RW)$ is 0.56 and 0.61 on the 2B--3B models, failing even the lenient
budget, and 0.31--0.47 at 7B--14B, so any wrong-answer budget below 0.31 defeats it at
every scale we test. A policy must read both axes. We do not claim that no scalar
combination of the two can work; we claim that neither single axis alone does.

\textbf{Factorized Abstention} answers iff $\scoreA \ge \thrA$ and $\scoreC \ge \thrC$, the
high-answerability, high-confidence corner of the geometry in
Figure~\ref{fig:geometry}. To make the guarantee cover the adaptive choice of thresholds,
we choose $(\thrA, \thrC)$ on a tuning split, freeze them, and certify on an independent
certification split with Clopper--Pearson upper bounds at confidence $1-\delta/2$ per
risk, so both hold simultaneously at $1-\delta = 0.90$ by Bonferroni; a held-out test
split gives empirical rates. The readout scores themselves are trained on the tuning
split only, so certification and test items are out-of-sample for scores and thresholds
alike (Appendix~\ref{app:policy}). The answerability score $\scoreA$ may be a hidden probe or the
elicited $P(\mathrm{IK})$; the policy is agnostic to its source. We calibrate and certify
the single-axis policies and a learned logistic combiner by the identical procedure. The
calibration machinery is standard split-conformal risk control in the Learn-Then-Test
family \citep{angelopoulos2021learn}; our contribution is the error decomposition and its
operational use, not the calibration method.

\begin{table}[t]
\centering
\caption{Test-split C-coverage of each policy at $\alpha_U{=}0.15$, $\alpha_W{=}0.50$,
$\delta{=}0.10$; \cmark{} marks policies whose both risk budgets certify on the
independent certification split; bold is the best certified coverage per model. All
Qwen2.5-3B policies fail certification: one accepted U item in 28 puts
$\mathrm{UCB}(\RU)$ at 0.16, and its answerability-only and combiner policies would
fail the W budget regardless (UCB 0.61). Exact $k/n$ counts, UCBs, per-split class
counts, and a target-sensitivity grid are in Appendix~\ref{app:policy}.}
\label{tab:policies}
\small
\resizebox{\linewidth}{!}{\begin{tabular}{lccccc}
\toprule
Policy (test C-coverage, cert.) & Gemma 2B ($n_C$=8) & Qwen 3B ($n_C$=8) & Qwen 7B ($n_C$=11) & Llama 8B ($n_C$=16) & Qwen 14B ($n_C$=17) \\
\midrule
Answer-confidence only & 0.12 \cmark & 0.38 & 0.27 \cmark & 0.31 \cmark & 0.41 \\
Answerability only & 0.50 & 0.12 & \textbf{0.55} \cmark & 0.44 \cmark & 0.35 \cmark \\
Learned joint scalar & \textbf{0.88} \cmark & 0.75 & 0.73 & \textbf{0.81} \cmark & 0.47 \\
Factorized & 0.62 & 0.75 & 1.00 & 0.75 \cmark & \textbf{0.47} \cmark \\
\bottomrule
\end{tabular}
}
\end{table}

The certificates expose an asymmetry a single scalar hides, and the asymmetry is
scale-dependent. The unanswerable-answer risk is certifiable almost everywhere: on the
independent certification splits the answerability gate accepts zero or one unanswerable
item, with one-sided upper bounds on $\RU$ of 0.07 to 0.16 (on Qwen2.5-3B the single
accepted U item puts the bound at 0.16, just missing the 0.15 budget, so no policy
certifies there). The wrong-answer risk is bounded by model accuracy. At 2B--3B, models
answer only 28 to 30 percent of answerable questions correctly, so holding $\RW$ forces
near-total abstention and the W budget certifies marginally or not at all. At 8B,
accuracy reaches 0.42, the ceiling lifts, and the factorized policy certifies both
budgets simultaneously at C-coverage 0.75, more than double the certified
answer-confidence-only coverage (0.31); at 14B the factorized policy is the only one
that certifies at all (C-coverage 0.47), because output confidence there is at its
blindest and both confidence-thresholding alternatives fail a budget. We state the
comparison against the strongest alternative plainly: the learned two-signal combiner
reaches the highest certified coverage wherever it certifies (0.88 on Gemma, 0.81 at
8B, versus 0.75 factorized at 8B), so if maximal coverage under these budgets is the
only goal, fusing the two signals into one scalar, as \citet{phillips2026entropy} do
for correctness, is the better policy where it certifies; at 14B it does not. What fusion cannot provide is the per-axis certificate: a deployment
that must report ``the unanswerable-answer rate is at most $x$'' separately from ``the
wrong-answer rate is at most $y$'' needs the factorized form, and the scientific claim
of this section is that both signals are required, not that the rectangular region is
optimal. Wrong-answer budgets tighter than $\alpha_W{=}0.5$ do not certify at any scale
we test, partly a certification-sample limit (18--22 W items per certification split):
the dual-risk decomposition makes explicit both what the model's capability supports and
what the calibration data can attest.

Two honesty notes on certification statistics. First, each certificate is a marginal
statement about one policy at one pre-declared target pair; the sensitivity grid in
Appendix~\ref{app:policy} is exploratory, and selecting after the fact the cell that
certified would void the guarantee (family-wise control over a target grid is exactly
the Learn-Then-Test extension we do not run here). Second, at these split sizes the
tuned thresholds are noisy: test coverage is not monotone in the budget across the grid
(a symptom of discrete threshold search on small tuning splits), and a certificate
issued from a zero-count bound can be optimistic about a fresh split. We therefore audit
the whole pipeline by replication: re-running split, probe training, tuning, and
certification end to end over 100 random resplits per model
(Appendix~\ref{app:replication}). The audit supports the protocol and sharpens the
scale story: among issued certificates, the unanswerable budget is never violated on
held-out test in any replicate of any model, and the wrong-answer budget is violated in
0 to 5.6 percent of issued replicates (at most 1 of 18), consistent with the 5 percent
per-bound allowance at these counts; certificates
issue in 18 of 100 resplits on Gemma 2 2B but 68 of 100 on Llama-3.1-8B, so at 8B
certification is not just possible but repeatable.

\section{Corroboration with surface-controlled, certified gates}
\label{sec:gates}

Because SelfAware answerability is surface-confounded, the cleaner evidence that an
internal answerability signal supports calibrated abstention comes from matched gates
that hold question form fixed and vary only the entity, under one frozen protocol with
group-level Clopper--Pearson selective-risk certificates. The settings are three matched
factual relations read by two model families (five model-relation settings); chemical
symbols and low-prominence birthplaces give the same answerability pattern in earlier
runs.

\begin{table}[t]
\centering
\caption{Matched factual-relation gates: internal vs output unknown-detection AUROC and
full three-way certification status.}
\label{tab:gates}
\small
\resizebox{\linewidth}{!}{%
\begin{tabular}{llccc}
\toprule
Relation & Reader (model) & groups & unk.\ AUROC (int.\ / out.) & full three-way \\
\midrule
Capitals  & Gemma Scope SAE (Gemma 2 2B-it) & 59 & 0.978 / 0.552 & certified (pooled) \\
Languages & Gemma Scope SAE (Gemma 2 2B-it) & 28 & 0.931 / 0.418 & certified (pooled) \\
Currencies & Gemma Scope SAE (Gemma 2 2B-it) & 28 & 1.000 / 0.549 & no (weak truth axis) \\
Capitals  & hidden probe (Qwen2.5-3B) & 59 & 0.802 / 0.522 & discrimination only \\
Languages & hidden probe (Qwen2.5-3B) & 28 & 0.870 / 0.452 & discrimination only \\
\bottomrule
\end{tabular}}
\end{table}

In every setting the internal answerability reader beats output-only uncertainty on
unknown detection, and on a population-matched subset of the capitals gate ($n{=}30$
groups), where an a-priori prominence proxy is at chance, the internal signal still
reaches AUROC 0.96, evidence that it is not reducible to entity prominence. Pooling two
relations (capitals and official languages on Gemma) certifies a full
true/false/unknown decision at a group-level selective-risk upper bound of 0.074 (one
error group in 51 answered groups, one-sided 90\%; 0.090 at 95\%), below a 10\%
target. Two caveats. The pooled pair was selected because the model's truth axis on
currencies proved unreliable in the same evaluation data, so the pooled certificate is
conditional on that post-hoc selection and we do not headline it. And these gates use a
different reader (a sparse-autoencoder feature battery for Gemma, selected on
calibration groups only, grounding the answerability predicate in named sparse features
in the spirit of conceptual grounding, \citealp{wagner2021interactive}) and a
group-level exchangeable unit; the full construction,
feature-selection protocol, prominence proxy, and split structure are documented in
Appendix~\ref{app:gates}. Every number in this paper is regenerated from committed
score artifacts and asserted by a continuous-integration script.

\section{Limitations}
\label{sec:limitations}

The geometry now spans four instruction-tuned models from three families, 2B to 8B, but
all are small by frontier standards; larger models may have better explicit
self-assessment, so we do not assert that the pattern persists at 70B and beyond,
and the hidden correctness readout already grows sharply from 3B to 8B. The 7B and 8B
models run 8-bit quantized (validated against full precision at 3B,
Appendix~\ref{app:compute}). SelfAware labels semantic answerability, whether a
determinate answer exists, not whether a particular model is competent to answer; we use
the term answerability throughout and treat model-relative competence separately in
Section~\ref{sec:gates}. CREPE contributes only the admissibility axis, because its
long-form ELI5 answers admit no alias-based correctness label, and its presupposition
annotations carry some inherent ambiguity. The answerability signal on SelfAware is
largely surface-recoverable (bag-of-words 0.87) and, model-dependently,
elicitation-recoverable ($P(\mathrm{IK})$ 0.61--0.88), so our claim about output is
restricted to ordinary answer-confidence there; on CREPE no elicitation we test
recovers it. Correctness is judged by alias matching on a small number of correct
answers and is read from one output statistic; richer correctness readers and a human
audit of alias matches would strengthen it. Our output baselines are single-pass
(confidence readouts, $P(\mathrm{IK})$, $P(\mathrm{True})$, premise-check);
sampling-based semantic entropy is not evaluated under this protocol. Our evidence is
observational plus policy: unlike \citet{slobodkin2023curious} and
\citet{lavi2026detecting} we do not intervene causally on the answerability direction. Certification splits contain only 18--22
wrong-answerable items per model, so wrong-answer budgets tighter than 0.5 cannot
certify regardless of signal quality; scaling the calibration data is the direct
remedy. All data are English.

\section{Conclusion}

Abstention is not one problem. Whether an attempted answer succeeded is visible in
ordinary answer-confidence; whether a question is admissible is not, although a hidden
probe recovers much of it, and, on some benchmarks and models, explicit elicitation or
surface cues recover part. Correct-answerable, wrong-answerable, and unanswerable questions occupy a
two-signal geometry that is stable from 2B to 8B across three model families, persists
on naturally occurring false presuppositions where trained output readouts sit at or
near chance (0.50--0.54), and pays off operationally: decomposing the two axes, rather than
thresholding one confidence scalar, yields better certifiable abstention, and the
dual-risk certificate itself measures what a given model's capability can and cannot
attest, with the wrong-answer budget becoming certifiable at useful coverage as models
grow more accurate. The contribution is not that hidden states know answerability, which
is established, but that answer correctness and question answerability induce
complementary score geometries on the same decisions, and that using both axes beats
using one. The thresholded conjunction of two probe-grounded predicates is the simplest
policy in a larger space: richer logical compositions over named, grounded predicates,
with the open-world boundary made explicit rather than implicit in a confidence
threshold \citep{wagner2022neural,badreddine2022logic}, are the natural next step, and
the repair result of Section~\ref{sec:repair} suggests the payoff extends beyond
abstention to routing among behaviors.

\subsubsection*{Ethics statement}
This work uses two public research benchmarks: SelfAware (questions curated by
\citealp{yin2023large}) and CREPE (questions from public Reddit ELI5 threads, curated
and annotated by \citealp{yu2023crepe}, BSD-licensed); we use question text and labels
only, no usernames or personal data. No human subjects were involved. Better abstention
mechanisms reduce the risk of models asserting false or fabricated answers; we do not
foresee direct negative applications, though certified risk bounds hold only under the
exchangeability assumptions stated and should not be quoted without them.

\subsubsection*{Reproducibility statement}
Every number in the paper is regenerated from committed score artifacts by deterministic
analysis scripts (fixed seeds) and asserted by a verification script run in continuous
integration; the scoring, analysis, and verification code and all datasets or their
loaders will be released. 

\bibliography{references}

@article{farquhar2024detecting,
  title   = {Detecting hallucinations in large language models using semantic entropy},
  author  = {Farquhar, Sebastian and Kossen, Jannik and Kuhn, Lorenz and Gal, Yarin},
  journal = {Nature},
  volume  = {630},
  number  = {8017},
  pages   = {625--630},
  year    = {2024},
  doi     = {10.1038/s41586-024-07421-0}
}

@article{yadkori2024mitigating,
  title   = {Mitigating {LLM} Hallucinations via Conformal Abstention},
  author  = {Abbasi Yadkori, Yasin and Kuzborskij, Ilja and Stutz, David and Gy{\"o}rgy, Andr{\'a}s and Fisch, Adam and Doucet, Arnaud and Beloshapka, Iuliya and Weng, Wei-Hung and Yang, Yao-Yuan and Szepesv{\'a}ri, Csaba and Cemgil, Ali Taylan and Tomasev, Nenad},
  journal = {arXiv preprint arXiv:2405.01563},
  year    = {2024}
}

@inproceedings{hendrycks2017baseline,
  title     = {A Baseline for Detecting Misclassified and Out-of-Distribution Examples in Neural Networks},
  author    = {Hendrycks, Dan and Gimpel, Kevin},
  booktitle = {International Conference on Learning Representations (ICLR)},
  year      = {2017}
}

@inproceedings{yin2023large,
  title     = {Do Large Language Models Know What They Don't Know?},
  author    = {Yin, Zhangyue and Sun, Qiushi and Guo, Qipeng and Wu, Jiawen and Qiu, Xipeng and Huang, Xuanjing},
  booktitle = {Findings of the Association for Computational Linguistics: ACL 2023},
  pages     = {8653--8665},
  year      = {2023}
}

@inproceedings{slobodkin2023curious,
  title     = {The Curious Case of Hallucinatory (Un)answerability: Finding Truths in the Hidden States of Over-Confident Large Language Models},
  author    = {Slobodkin, Aviv and Goldman, Omer and Caciularu, Avi and Dagan, Ido and Ravfogel, Shauli},
  booktitle = {Proceedings of the 2023 Conference on Empirical Methods in Natural Language Processing (EMNLP)},
  pages     = {3607--3625},
  year      = {2023}
}

@inproceedings{lavi2026detecting,
  title     = {Detecting (Un)answerability in Large Language Models with Linear Directions},
  author    = {Lavi, Maor Juliet and Milo, Tova and Geva, Mor},
  booktitle = {Proceedings of the 19th Conference of the European Chapter of the Association for Computational Linguistics (EACL), Volume 1: Long Papers},
  pages     = {682--699},
  year      = {2026}
}

@inproceedings{chen2026query,
  title     = {Query-Level Uncertainty in Large Language Models},
  author    = {Chen, Lihu and de Melo, Gerard and Suchanek, Fabian M. and Varoquaux, Ga{\"e}l},
  booktitle = {International Conference on Learning Representations (ICLR)},
  year      = {2026}
}

@article{kadavath2022language,
  title   = {Language Models (Mostly) Know What They Know},
  author  = {Kadavath, Saurav and Conerly, Tom and Askell, Amanda and Henighan, Tom and Drain, Dawn and Perez, Ethan and Schiefer, Nicholas and Hatfield-Dodds, Zac and DasSarma, Nova and Tran-Johnson, Eli and others},
  journal = {arXiv preprint arXiv:2207.05221},
  year    = {2022}
}

@article{phillips2026entropy,
  title   = {Entropy Alone is Insufficient for Safe Selective Prediction in {LLMs}},
  author  = {Phillips, Edward and Gustafsson, Fredrik K. and Wu, Sean and Thakur, Anshul and Clifton, David A.},
  journal = {arXiv preprint arXiv:2603.21172},
  year    = {2026}
}

@article{patel2026uncertainty,
  title   = {Are {LLM} Uncertainty and Correctness Encoded by the Same Features? A Functional Dissociation via Sparse Autoencoders},
  author  = {Patel, Het and Chen, Tianyi and Wei, Hua and Papalexakis, Evangelos E. and Chen, Jiayu},
  journal = {arXiv preprint arXiv:2604.19974},
  year    = {2026}
}

@article{angelopoulos2021learn,
  title   = {Learn then Test: Calibrating Predictive Algorithms to Achieve Risk Control},
  author  = {Angelopoulos, Anastasios N. and Bates, Stephen and Cand{\`e}s, Emmanuel J. and Jordan, Michael I. and Lei, Lihua},
  journal = {The Annals of Applied Statistics},
  volume  = {19},
  number  = {2},
  pages   = {1641--1662},
  year    = {2025},
  note    = {arXiv:2110.01052}
}

@inproceedings{mohri2024language,
  title     = {Language Models with Conformal Factuality Guarantees},
  author    = {Mohri, Christopher and Hashimoto, Tatsunori},
  booktitle = {Proceedings of the 41st International Conference on Machine Learning (ICML)},
  year      = {2024}
}

@inproceedings{azaria2023internal,
  title     = {The Internal State of an {LLM} Knows When It's Lying},
  author    = {Azaria, Amos and Mitchell, Tom},
  booktitle = {Findings of the Association for Computational Linguistics: EMNLP 2023},
  pages     = {967--976},
  year      = {2023}
}

@inproceedings{chen2024inside,
  title     = {{INSIDE}: {LLMs}' Internal States Retain the Power of Hallucination Detection},
  author    = {Chen, Chao and Liu, Kai and Chen, Ze and Gu, Yi and Wu, Yue and Tao, Mingyuan and Fu, Zhihang and Ye, Jieping},
  booktitle = {International Conference on Learning Representations (ICLR)},
  year      = {2024}
}

@article{kossen2024semantic,
  title   = {Semantic Entropy Probes: Robust and Cheap Hallucination Detection in {LLMs}},
  author  = {Kossen, Jannik and Han, Jiatong and Razzak, Muhammed and Schut, Lisa and Malik, Shreshth and Gal, Yarin},
  journal = {arXiv preprint arXiv:2406.15927},
  year    = {2024}
}

@inproceedings{mucsanyi2024benchmarking,
  title     = {Benchmarking Uncertainty Disentanglement: Specialized Uncertainties for Specialized Tasks},
  author    = {Mucs{\'a}nyi, B{\'a}lint and Kirchhof, Michael and Oh, Seong Joon},
  booktitle = {Advances in Neural Information Processing Systems (NeurIPS), Datasets and Benchmarks Track},
  year      = {2024}
}

@article{morenocencerrado2025noanswer,
  title   = {No Answer Needed: Predicting {LLM} Answer Accuracy from Question-Only Linear Probes},
  author  = {Moreno Cencerrado, Iv{\'a}n Vicente and Padr{\'e}s Masdemont, Arnau and Gonzalvez Hawthorne, Anton and Africa, David Demitri and Pacchiardi, Lorenzo},
  journal = {arXiv preprint arXiv:2509.10625},
  year    = {2025},
  note    = {ICLR 2026 Workshop on Principled Design for Trustworthy AI}
}

@inproceedings{yu2023crepe,
  title     = {{CREPE}: Open-Domain Question Answering with False Presuppositions},
  author    = {Yu, Xinyan and Min, Sewon and Zettlemoyer, Luke and Hajishirzi, Hannaneh},
  booktitle = {Proceedings of the 61st Annual Meeting of the Association for Computational Linguistics (ACL), Volume 1: Long Papers},
  pages     = {10457--10480},
  year      = {2023}
}

@inproceedings{huang2024large,
  title     = {Large Language Models Cannot Self-Correct Reasoning Yet},
  author    = {Huang, Jie and Chen, Xinyun and Mishra, Swaroop and Zheng, Huaixiu Steven and Yu, Adams Wei and Song, Xinying and Zhou, Denny},
  booktitle = {International Conference on Learning Representations (ICLR)},
  year      = {2024}
}

@inproceedings{wagner2022neural,
  title     = {Neural-Symbolic Reasoning Under Open-World and Closed-World Assumptions},
  author    = {Wagner, Benedikt and d'Avila Garcez, Artur},
  booktitle = {AAAI Spring Symposium on Machine Learning and Knowledge Engineering (AAAI-MAKE), CEUR Workshop Proceedings Vol-3121},
  year      = {2022}
}

@article{wagner2021interactive,
  title   = {Neural-Symbolic Integration for Interactive Learning and Conceptual Grounding},
  author  = {Wagner, Benedikt and d'Avila Garcez, Artur},
  journal = {arXiv preprint arXiv:2112.11805},
  year    = {2021}
}

@article{badreddine2022logic,
  title   = {Logic Tensor Networks},
  author  = {Badreddine, Samy and d'Avila Garcez, Artur and Serafini, Luciano and Spranger, Michael},
  journal = {Artificial Intelligence},
  volume  = {303},
  pages   = {103649},
  year    = {2022}
}
\bibliographystyle{iclr2026_conference}

\appendix

\section{Per-layer readouts and geometry details}
\label{app:layers}

This appendix collects the full per-layer readout values behind the nested layer
selection of Section~\ref{sec:geometry} (Table~\ref{tab:perlayer}), the complete
standardized mean gaps and within-signal ratios behind the ratio-asymmetry statement
(Table~\ref{tab:gaps}), and the label-artifact control in which the answerability
direction is trained on C-versus-U items only (Table~\ref{tab:cvu}).

\begin{table}[h]
\centering
\caption{Hidden-state readout AUROC by layer (fractional depths 0.4/0.6/0.8 of each
model). The answerability readout is near ceiling at every probed depth; the correctness
readout improves with depth and scale.}
\label{tab:perlayer}
\small
\begin{tabular}{lcccccc}
\toprule
Model & \multicolumn{3}{c}{H$\to$C by depth} & \multicolumn{3}{c}{H$\to$A by depth} \\
 & 0.4 & 0.6 & 0.8 & 0.4 & 0.6 & 0.8 \\
\midrule
Gemma 2 2B-it & 0.64 & 0.72 & 0.74 & 0.98 & 0.98 & 0.98 \\
Qwen2.5-3B & 0.59 & 0.59 & 0.58 & 0.97 & 0.97 & 0.98 \\
Qwen2.5-7B & 0.77 & 0.80 & 0.83 & 0.98 & 0.98 & 0.99 \\
Llama-3.1-8B & 0.79 & 0.85 & 0.88 & 0.99 & 0.99 & 0.99 \\
Qwen2.5-14B & 0.79 & 0.86 & 0.86 & 0.98 & 0.99 & 0.99 \\
\bottomrule
\end{tabular}

\end{table}

\begin{table}[h]
\centering
\caption{All standardized mean gaps behind the ratio asymmetry of
Section~\ref{sec:geometry}: C$-$W and W$-$U point differences per signal, and the
within-signal ratio of the smaller to the larger gap (ratios computed before rounding).}
\label{tab:gaps}
\small
\begin{tabular}{lcccccc}
\toprule
 & \multicolumn{3}{c}{output confidence} & \multicolumn{3}{c}{answerability readout} \\
Model & C$-$W & W$-$U & ratio & C$-$W & W$-$U & ratio \\
\midrule
Gemma 2 2B-it & 1.00 & -0.19 & 0.19 & 0.04 & 1.76 & 0.02 \\
Qwen2.5-3B & 0.93 & -0.32 & 0.35 & 0.01 & 1.83 & 0.00 \\
Qwen2.5-7B & 0.82 & 0.03 & 0.03 & 0.09 & 1.73 & 0.05 \\
Llama-3.1-8B & 1.03 & 0.14 & 0.14 & 0.12 & 1.67 & 0.07 \\
Qwen2.5-14B & 0.82 & -0.16 & 0.20 & 0.04 & 1.77 & 0.02 \\
\bottomrule
\end{tabular}

\end{table}

\begin{table}[h]
\centering
\caption{Label-artifact control: a direction trained on C-versus-U only (no W items in
training) still places held-out W with C, not U.}
\label{tab:cvu}
\small
\begin{tabular}{lcccc}
\toprule
Model & W vs.\ U AUROC & C vs.\ W AUROC & W position (U=0, C=1) & $n_W$ \\
\midrule
Gemma 2 2B-it & 0.97 & 0.60 & 0.85 & 54 \\
Qwen2.5-3B & 0.97 & 0.62 & 0.84 & 36 \\
Qwen2.5-7B & 0.98 & 0.66 & 0.81 & 51 \\
Llama-3.1-8B & 0.98 & 0.70 & 0.78 & 42 \\
Qwen2.5-14B & 0.98 & 0.70 & 0.71 & 46 \\
\bottomrule
\end{tabular}

\end{table}

\FloatBarrier
\section{Policy certification details}
\label{app:policy}

The readout scores are trained on the tuning split (which coincides with the analysis
calibration split), so the certification and test splits are out-of-sample for both the
scores and the thresholds; thresholds are chosen on the tuning split by maximizing
tune-split C-coverage subject to tune-split feasibility and then frozen. Certification
uses exact one-sided binomial (Clopper--Pearson) upper bounds at level $\delta/2$ per
risk, giving simultaneous $1-\delta$ coverage by Bonferroni.

\begin{table}[h]
\centering
\caption{Full certification details at $\alpha_U{=}0.15$, $\alpha_W{=}0.50$,
$\delta{=}0.10$: accepted counts and Clopper--Pearson upper bounds per risk on the
certification split; C-coverage on the held-out test split.}
\label{tab:policyapp}
\small
\begin{tabular}{llcccc}
\toprule
Model & Policy & C-cov & $\RU$ $k/n$, UCB & $\RW$ $k/n$, UCB & cert.\ \\
\midrule
Gemma 2 2B-it & Answer-conf.\ only & 0.12 & 0/41, 0.07 & 0/22, 0.13 & yes \\
 & Answerability only & 0.50 & 0/41, 0.07 & 8/22, 0.56 & no \\
 & Learned joint & 0.88 & 0/41, 0.07 & 6/22, 0.47 & yes \\
 & Factorized & 0.62 & 0/41, 0.07 & 7/22, 0.52 & no \\
\midrule
Qwen2.5-3B & Answer-conf.\ only & 0.38 & 1/28, 0.16 & 0/18, 0.15 & no \\
 & Answerability only & 0.12 & 1/28, 0.16 & 7/18, 0.61 & no \\
 & Learned joint & 0.75 & 1/28, 0.16 & 7/18, 0.61 & no \\
 & Factorized & 0.75 & 1/28, 0.16 & 5/18, 0.50 & no \\
\midrule
Qwen2.5-7B & Answer-conf.\ only & 0.27 & 2/41, 0.15 & 0/22, 0.13 & yes \\
 & Answerability only & 0.55 & 0/41, 0.07 & 6/22, 0.47 & yes \\
 & Learned joint & 0.73 & 0/41, 0.07 & 8/22, 0.56 & no \\
 & Factorized & 1.00 & 1/41, 0.11 & 9/22, 0.60 & no \\
\midrule
Llama-3.1-8B & Answer-conf.\ only & 0.31 & 0/41, 0.07 & 1/18, 0.24 & yes \\
 & Answerability only & 0.44 & 0/41, 0.07 & 2/18, 0.31 & yes \\
 & Learned joint & 0.81 & 0/41, 0.07 & 3/18, 0.38 & yes \\
 & Factorized & 0.75 & 0/41, 0.07 & 3/18, 0.38 & yes \\
\midrule
Qwen2.5-14B & Answer-conf.\ only & 0.41 & 3/41, 0.18 & 1/23, 0.19 & no \\
 & Answerability only & 0.35 & 0/41, 0.07 & 6/23, 0.45 & yes \\
 & Learned joint & 0.47 & 0/41, 0.07 & 9/23, 0.58 & no \\
 & Factorized & 0.47 & 0/41, 0.07 & 7/23, 0.50 & yes \\
\bottomrule
\end{tabular}

\end{table}

\begin{table}[h]
\centering
\caption{Target-sensitivity grid for the factorized policy: test C-coverage and
certification status across $(\alpha_U, \alpha_W)$; \cmark{} = both budgets certified,
-- = not certified.}
\label{tab:sens}
\small
\begin{tabular}{llccc}
\toprule
Model & $\alpha_U$ & $\alpha_W{=}0.20$ & $\alpha_W{=}0.30$ & $\alpha_W{=}0.50$ \\
\midrule
Gemma 2 2B-it & 0.10 & 0.75 \xmark & 0.62 \xmark & 0.62 \xmark \\
 & 0.15 & 0.75 \xmark & 0.62 \xmark & 0.62 \xmark \\
 & 0.20 & 0.75 \xmark & 0.62 \xmark & 0.62 \xmark \\
\midrule
Qwen2.5-3B & 0.10 & 0.50 \xmark & 0.62 \xmark & 0.75 \xmark \\
 & 0.15 & 0.50 \xmark & 0.62 \xmark & 0.75 \xmark \\
 & 0.20 & 0.50 \xmark & 0.62 \xmark & 0.75 \cmark \\
\midrule
Qwen2.5-7B & 0.10 & 0.55 \xmark & 0.64 \xmark & 1.00 \xmark \\
 & 0.15 & 0.55 \xmark & 0.64 \xmark & 1.00 \xmark \\
 & 0.20 & 0.55 \xmark & 0.64 \xmark & 1.00 \xmark \\
\midrule
Llama-3.1-8B & 0.10 & 0.31 \xmark & 0.56 \xmark & 0.75 \cmark \\
 & 0.15 & 0.31 \xmark & 0.56 \xmark & 0.75 \cmark \\
 & 0.20 & 0.31 \xmark & 0.56 \xmark & 0.75 \cmark \\
\midrule
Qwen2.5-14B & 0.10 & 0.53 \xmark & 0.53 \xmark & 0.47 \cmark \\
 & 0.15 & 0.53 \xmark & 0.53 \xmark & 0.47 \cmark \\
 & 0.20 & 0.53 \xmark & 0.53 \xmark & 0.47 \cmark \\
\bottomrule
\end{tabular}

\end{table}

\begin{table}[h]
\centering
\caption{CREPE certified admissibility gate ($\alpha_{FP}{=}0.15$, $\delta{=}0.05$,
leak-free four-way split): false-premise answer rate on the certification and test
splits at the frozen threshold, with coverage of admissible questions.}
\label{tab:crepegate}
\small
\begin{tabular}{lccccc}
\toprule
Model & FP-rate cert $k/n$, UCB & cov.\ (cert) & FP-rate test $k/n$ & cov.\ (test) & cert.\ \\
\midrule
Gemma 2 2B-it & 3/77, 0.098 & 0.07 & 6/91 (0.066) & 0.13 & yes \\
Qwen2.5-3B & 5/77, 0.132 & 0.17 & 8/91 (0.088) & 0.09 & yes \\
Qwen2.5-7B & 5/77, 0.132 & 0.23 & 8/91 (0.088) & 0.23 & yes \\
Llama-3.1-8B & 0/77, 0.038 & 0.14 & 7/91 (0.077) & 0.25 & yes \\
Qwen2.5-14B & 8/77, 0.180 & 0.24 & 6/91 (0.066) & 0.27 & no \\
\bottomrule
\end{tabular}

\end{table}

\section{Prompts and datasets}
\label{app:prompts}
All scoring uses a single user turn: ``Answer the question in a few words.\ Question:
\{q\}'' with each model's chat template and greedy decoding (24 new tokens maximum). The
$P(\mathrm{IK})$ elicitation asks ``Can you correctly answer the following question?
Answer yes or no.''\ and reads the renormalized probability of yes; the premise-check
elicitation asks ``Does the following question rest on a false or mistaken premise?
Answer yes or no.''\ and reads the probability of no. SelfAware samples are balanced per
class with a fixed seed: 150 answerable and 150 unanswerable per model (120 each for
Qwen2.5-3B). CREPE samples are 500 \emph{normal} and 500 \emph{false presupposition}
questions per model, drawn with a fixed seed from the official test split (3{,}004
questions, 751 with false presuppositions), restricted to questions of at most 300
characters (applied identically to both classes). Probes are $L_2$-regularized logistic
readouts on the final prompt-token hidden state; all splits, seeds, and thresholds are
fixed in the released code, and a verification script asserts every number in this paper
against its committed score artifact.

\section{Certificate replication audit}
\label{app:replication}

For each dataset and model we re-run the complete certification pipeline (random
resplit, probe retraining, threshold tuning, certification, held-out test) over 100
seeds. A certificate is \emph{issued} when a tune-feasible threshold exists and the
frozen threshold passes its certification-split bound; an issued certificate is
\emph{violated} when the held-out test split's empirical rate exceeds the target. If
the protocol is sound, violations among issued certificates should occur in at most
about a $\delta$ fraction of replicates.

\begin{table}[h]
\centering
\caption{Replication audit over 100 resplits per model: certificates issued, test-split
violation rate among issued certificates (for the factorized policy, the $\RU$ and
$\RW$ rates are shown separately), and coverage distribution (mean $\pm$ sd).}
\label{tab:replication}
\small
\begin{tabular}{llccc}
\toprule
Protocol & Model & issued & violation rate & coverage \\
\midrule
CREPE gate & Gemma 2 2B-it & 55/100 & 0.00 & 0.23 $\pm$ 0.07 \\
CREPE gate & Qwen2.5-3B & 53/100 & 0.00 & 0.18 $\pm$ 0.06 \\
CREPE gate & Qwen2.5-7B & 45/100 & 0.00 & 0.23 $\pm$ 0.07 \\
CREPE gate & Llama-3.1-8B & 53/100 & 0.04 & 0.23 $\pm$ 0.08 \\
\midrule
Factorized (SelfAware) & Gemma 2 2B-it & 18/100 & 0.00 / 0.06 & 0.65 $\pm$ 0.16 \\
Factorized (SelfAware) & Qwen2.5-3B & 32/100 & 0.00 / 0.00 & 0.55 $\pm$ 0.14 \\
Factorized (SelfAware) & Qwen2.5-7B & 22/100 & 0.00 / 0.04 & 0.70 $\pm$ 0.16 \\
Factorized (SelfAware) & Llama-3.1-8B & 68/100 & 0.00 / 0.00 & 0.59 $\pm$ 0.16 \\
\bottomrule
\end{tabular}

\end{table}

\section{Premise-routing pilot details and replication}
\label{app:router}

Pilot protocol: 120 false-premise and 120 sound questions per model from the held-out
(non-probe-train) half of the CREPE sample; standard-prompt generations reused from the
committed scoring runs; challenge-prompt generations produced with the same greedy
decoding (48 token maximum). The strict premise-contest detector requires an explicit
premise-denial template; the marker set and all generations are released for audit.
Hypotheses and kill-criteria were fixed in the script header before running.

\section{Cross-model control values}
\label{app:controls}

Table~\ref{tab:controls} reports the output-side elicitation controls on SelfAware for
every model. The spread of the $P(\mathrm{IK})$ column (0.59 to 0.88) is the
model-dependence discussed in Section~\ref{sec:controls}: whether explicit
self-assessment recovers answerability depends on the model, and no elicitation matches
the hidden readout on any of them.

\begin{table}[h]
\centering
\caption{Output-side elicitation controls on SelfAware per model: $P(\mathrm{IK})$
(pre-generation self-assessment) and $P(\mathrm{True})$ (post-generation
self-evaluation of the produced answer) as answerability readers, alongside the trained
output readout (O$\to$A) and the hidden readout (H$\to$A) from
Table~\ref{tab:2x2}.}
\label{tab:controls}
\small
\begin{tabular}{lcccc}
\toprule
Model & O$\to$A readout & $P(\mathrm{IK})$ & $P(\mathrm{True})$ & H$\to$A readout \\
\midrule
Gemma 2 2B-it & 0.56 & 0.88 & 0.78 & 0.98 \\
Qwen2.5-3B & 0.62 & 0.61 & 0.65 & 0.97 \\
Qwen2.5-7B & 0.64 & 0.66 & 0.69 & 0.98 \\
Llama-3.1-8B & 0.67 & 0.81 & 0.62 & 0.99 \\
Qwen2.5-14B & 0.54 & 0.59 & 0.65 & 0.99 \\
\bottomrule
\end{tabular}

\end{table}

\FloatBarrier
\section{Matched-gate construction and SAE reader protocol}
\label{app:gates}

\textbf{Gates.} Each gate is a set of matched groups for one factual relation (capital,
official language, currency). A group contains a TRUE item (real entity, correct
object), a FALSE item (same entity, wrong but type-consistent object), and an UNKNOWN
item (a real, lexically ordinary entity outside the model's reliable competence, e.g.\
Nauru or Comoros for capitals). Prominence tiers are assigned a priori by a documented
source-side proxy (fixed lists / \texttt{wikibase:sitelinks}), before any model
scoring; answering an UNKNOWN item counts as an answerability error regardless of
string-level truth. The split unit is the group (fixed seed); probe fitting, SAE feature
selection, and threshold selection use calibration groups only, and all reported numbers
are computed on test groups.

\textbf{SAE reader.} For Gemma 2 2B-it we use Gemma~Scope residual-stream sparse
autoencoders (\texttt{gemma-scope-2b-pt-res}, width 16k), with two candidate depths
(layers 10 and 14) and the depth chosen on calibration groups. The answerability
support score is a normalized battery of the top $k{=}16$ SAE features ranked on
calibration groups only; normalization bounds are also calibration-fitted. The Qwen2.5-3B
reader is an $L_2$-regularized logistic probe on the final prompt-token hidden state,
fit on calibration groups. Risk control uses a fixed 200-point threshold grid with
selective-risk target $\alpha{=}0.10$, $\delta{=}0.10$, thresholds selected on
calibration groups and evaluated on test groups; group-level certificates use exact
binomial bounds on per-group answered/error indicators, with the group as the
exchangeable unit.

\section{Compute and quantization}
\label{app:compute}
All experiments run on a single Apple M4 laptop (24\,GB unified memory). The 2B and 3B
models are scored in full precision with PyTorch; the 7B and 8B models are scored with
8-bit quantized weights via MLX. The quantized pipeline is validated against the
full-precision pipeline on Qwen2.5-3B: identical items and protocol give identical
answerable accuracy (0.300) and readout AUROCs within bootstrap noise (H$\to$A 0.972 vs
0.973; O$\to$A 0.611 vs 0.625). Scoring a 1{,}000-item dataset takes 15--30 minutes per
7--8B model.

\end{document}